\documentclass[conference,letterpaper]{IEEEtran}
\IEEEoverridecommandlockouts
\usepackage{cite}
\usepackage{amsmath,amssymb,amsfonts}
\usepackage{graphicx}
\usepackage{textcomp}
\usepackage{xcolor}
\def\BibTeX{{\rm B\kern-.05em{\sc i\kern-.025em b}\kern-.08em
    T\kern-.1667em\lower.7ex\hbox{E}\kern-.125emX}}

\usepackage{booktabs}
\usepackage{multirow}
\usepackage{longtable}

\usepackage{dblfloatfix} 

\let\labelindent\relax
\usepackage{enumitem}
\usepackage{algorithm}
\usepackage[noend]{algpseudocode}
\usepackage{lipsum}
\makeatletter
\renewcommand{\ALG@beginalgorithmic}{\small}
\makeatother

\begin{document}

\title{On the Performance Analysis of the Adversarial System Variant Approximation Method to Quantify Process Model Generalization}

\author{{Julian~Theis, Ilia~Mokhtarian, and~Houshang~Darabi}
\thanks{J. Theis, I. Mokhtarian, and H. Darabi are with University of Illinois at Chicago, Department of Mechanical and Industrial Engineering, 842 West Taylor Street, Chicago, IL 60607, United States. H. Darabi is the corresponding author. E-mail: \{jtheis3, imokht2, hdarabi\}@uic.edu.}}

\maketitle
\begin{abstract}
Process mining algorithms discover a process model from an event log. The resulting process model is supposed to describe all possible event sequences of the underlying system. Generalization is a process model quality dimension of interest. A generalization metric should quantify the extent to which a process model represents the observed event sequences contained in the event log and the unobserved event sequences of the system. Most of the available metrics in the literature cannot properly quantify the generalization of a process model. A recently published method \cite{theis2020adversarial} called Adversarial System Variant Approximation leverages Generative Adversarial Networks to approximate the underlying event sequence distribution of a system from an event log. While this method demonstrated performance gains over existing methods in measuring the generalization of process models, its experimental evaluations have been performed under ideal conditions. This paper experimentally investigates the performance of Adversarial System Variant Approximation under non-ideal conditions such as biased and limited event logs. Moreover, experiments are performed to investigate the originally proposed sampling hyperparameter value of the method on its performance to measure the generalization. The results confirm the need to raise awareness about the working conditions of the Adversarial System Variant Approximation method. The outcomes of this paper also serve to initiate future research directions.
\end{abstract}


\section{Introduction}\label{sec:introduction}
In recent years, process mining has developed into a major research discipline. Significant research effort has been spent on the automated discovery of process models from event logs, i.e. \textit{process discovery}, and the quality assessment of such models, called \textit{conformance checking}. 
While the focus of conformance checking has been mainly on measuring how well a discovered process model reflects event sequences that are recorded in an event log, measuring the extent to which a process model generalizes the possible event sequences of the system, from which the event log originates, has been neglected. 
The origin of such event logs are usually real-world systems that are applied in domains such as e.g. business processes \cite{van2007business}, manufacturing \cite{yang2014system, Theis2019c} or healthcare \cite{ghasemi2016process, darabi2009hospital, diabetesicu}.
Research studies have shown that measuring the generalization of discovered process models is of importance \cite{Rehse2018} and that only a few proposed methods focus on this objective. At the same time, the research community is aware that the existing methods do not fully address the required needs and present individual major shortcomings \cite{Syring2019, Janssenswillen2018}.

Adversarial System Variant Approximation (AVATAR) is a novel methodology \cite{theis2020adversarial} to overcome some of the known issues in measuring generalization. 
This method leverages a Generative Adversarial Network (GAN) to train a neural network on the same event log that is used to discover a process model. 
AVATAR is based on the fact that GANs successfully demonstrated the ability to unveil underlying data distributions in domains such as computer vision \cite{goodfellow2014generative} and transfers the approach to the context of measuring the generalization of process models. 
By sampling from the GAN that is trained on the event log, a baseline of supposedly generalizing event sequences is obtained. 
Experimental evaluations have been performed using ground truth systems. 
These evaluations have shown that the GAN of AVATAR can model observed event sequences that are contained in the event log and unobserved event sequences of the ground truth system more accurately than state-of-the-art process discovery algorithms. 

Whereas the experimental evaluation of AVATAR demonstrated that GANs are suitable and promising neural network architectures that can be used to measure the generalization of a process model, further research is required to understand the working conditions of those GANs in depth. 
This is of utmost importance to improve and mature the AVATAR methodology. 
This paper contributes to this objective by conducting performance analyses on the GANs of AVATAR using the same ground truth systems that were used in the original publication. 
First, the performance analyses include an experimental evaluation of the proposed sampling hyperparameter value $k$ of $10,000$ of the GAN of AVATAR since a comprehensive analysis remains unpublished to date.
Second, experiments are performed on limited event log sizes. The original publication used a constant 70\% split ratio of the event sequences of the ground truth systems that were used as the event log for process discovery and AVATAR. Under real-world conditions, such a constant 70\% split ratio is usually infeasable. Hence, it is necessary to investigate the GAN performance of AVATAR with different split ratios.
Third, an experimental evaluation is performed on the robustness of AVATAR towards bias. Specifically, this paper investigates if event logs that are biased affect the ability of the GAN to unveil unobserved event sequences from the ground truth system.
The results of the experiments are used to draw conclusions and to raise awareness about the working conditions of the GANs of AVATAR. This in turn provides future research directions to mature the AVATAR methodology.

The paper is structured as follows. Related work including an introduction of the AVATAR methodology is described in Section \ref{sec:related-work}, followed by an introduction of notations that are used throughout the paper in Section \ref{sec:notations}. The research questions of this paper are elaborated in Section \ref{sec:problem-statement}. Section \ref{sec:exp} introduces the experimental setups,  followed by a discussion of the obtained results in Section \ref{sec:results}. Section \ref{sec:conclusion} concludes the paper and provides future research directions.

\section{Related Work}\label{sec:related-work}
The content of this section is two-fold. First, a brief introduction to generalization metrics and their historical development is provided in Section \ref{sec:gen-metric}. Second, an introduction of the novel AVATAR method is provided in Section \ref{sec:avatar}.

\subsection{Generalization Metric}\label{sec:gen-metric}
Generalization describes that a process model, such as a Petri net, models ideally all possible event sequences of a system that can realistically occur. 
This means that a process model should allow for the event sequences that are recorded in an event log when observing a system under investigation. These event sequences are usually used to automatically discover a process model using a process discovery algorithm. 
Additionally, the process model should not allow for unrealistic event sequences beyond the observed ones. It is obvious that the difficulty of measuring the generalization of a process model reduces to classifying if given unobserved event sequences are either realistic or unrealistic in the context of the system under investigation.

A significant amount of research has been spent on measuring how well a process model allows for event sequences contained in an event log (i.e. measuring the \textit{fitness}) and how well a process model restricts to allow for event sequences beyond the ones contained in an event log (i.e. measuring the \textit{precision}). However, research on measuring the generalization of process models is scarce due to the difficulty of deriving realistic and unobserved event sequences from an event log.
Nonetheless, the process mining research community is aware that the quality dimension of generalization is of importance \cite{generalization, Syring2019, Janssenswillen2018}.

Historically, one of the first approaches to quantify the extent to which a process model generalizes event sequences beyond the ones contained in an event log has been introduced by Buijs et al. \cite{buijs2012role}. The proposed approach is based on quantifying the trustworthiness of the precision of a process model using alignments. Highly frequent used areas of a process model are considered well generalizing whereas low frequent parts of the model are less generalizing. 

Van der Aalst et al. \cite{replay2012} built a measurement to quantify that a process model does not overfit on a given event log. Specifically, their approach is based on the probability of observing a new event in any given state of the model based on the observations contained in the event log. If the likelihood of observing a new event in a given state is small, then the generalization is good.

Brouck et al. \cite{vanden2013determining} introduced a method to measure the generalization of a process model using weighted artificial negative events. In comparison to an actual event, an artificial negative event prevents the occurrence of a specific event at a given time. This concept enables to derive allowed and disallowed generalized event sequences.

A further method has been proposed by van Dongen et al. \cite{van2016unified}. This method is based on anti-alignments which are event sequences that are disparate from a set of given event sequences. This notion is used to measure the generalization by relating the state space of a process model. A generalizing process model has therefore a maximally different set of anti-alignments without introducing unseen states.

A comparative study has been conducted by Janssenswillen et al. \cite{janssenswillen2017comparative}. Their investigations lead to the conclusion that metrics that quantify the generalization with respect to a given event log do usually not assess the quality of a process model concerning the underlying system correctly. Hence, generalization metrics need to be developed that do not solely relate modeled event sequences to the sequences contained in an event log. Such metrics should be evaluated using ground truth systems. 

\subsection{Adversarial System Variant Approximation}\label{sec:avatar}
AVATAR is a recently proposed approach to quantify the extent to which a process model generalizes \cite{theis2020adversarial}.
The idea of this method is to unveil realistic but unobserved event sequences of a system using Generative Adversarial Networks (GANs) \cite{goodfellow2014generative}. If it is possible to confidently model unobserved event sequences using GANs, then measuring the generalization reduces to measuring the fitness and precision of a process model using the observed event log in combination with the unobserved event sequences that are modeled by the GAN. This is motivated by the generalization capabilities of GANs \cite{Arora2017}.

\begin{figure}[h]
  \begin{center}
    \includegraphics[width=230pt]{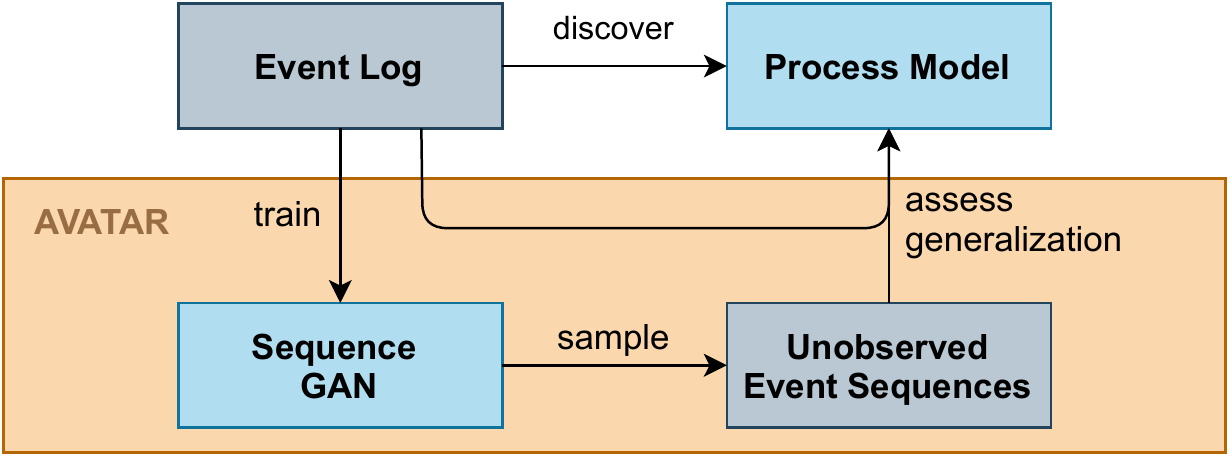}
  \end{center}
  \caption{Flow chat of the AVATAR methodology, derived from \cite{theis2020adversarial}}
  \label{fig:avatar_flowchart}
\end{figure}

A flow chart of the methodology is provided in Figure \ref{fig:avatar_flowchart}. A given set of event sequences that is used for automated process discovery is also used to train a Sequence GAN. AVATAR leverages a RelGAN \cite{nie2018relgan} architecture that is enhanced with an additional standard discriminator neural network. A major hyperparameter of this Sequence GAN architecture is the temperature control $\beta$ of the RelGAN that controls the tradeoff between sample diversity and quality. The trained Sequence GAN is then used to sample unobserved event sequences. AVATAR proposes therefore two sampling methodologies. 
The first is  \textit{naive sampling} controlled by the hyperparameter $k$ which means that $k$ samples are drawn from the generator of the Sequence GAN. The intuition is that the number of unique event sequences converges with an increasing number of sampling iterations. This also means that the relative frequency of an event sequence indicates the modeling confidence of this particular event sequence.
The second sampling methodology uses the \textit{Metropolis-Hastings algorithm} \cite{metropolis1953equation} and is inspired by the work of Turner et al. \cite{Turner2018}.
It is assumed that by sampling from the Sequence GAN, the unobserved event sequences of a system can be unveiled. In this case, quantifying the generalization of a process model reduces to measuring the fitness and precision of the process model with respect to the set of observed and approximated unobserved event sequences from the GAN.

The AVATAR methodology has been statistically evaluated using the finite set of event sequences of $15$ ground truth Petri nets. These Petri nets were created artificially as part of a comparative study of process discovery quality measures \cite{janssenswillen2017comparative} and are publicly available\footnote{https://github.com/gertjanssenswillen/processquality/}. Each of the $15$ Petri nets has different characteristics. $10$ of the Petri nets can be classified as \textit{moderate complex} with a small number of transitions and comparatively few parallelisms whereas $5$ Petri nets are \textit{highly complex} with a larger number of transitions and parallel structures. The \textit{highly complex} Petri nets are supposed to reflect the complexity of real-world systems. 
For each ground truth Petri net, a random and unbiased 70\% random split of the modeled unique event sequences was considered as an event log. These event logs were used to discover process models using two process discovery algorithms \cite{vanden2017fodina, augusto2019split}.
The remaining 30\% were withheld as the set of unobserved event sequences that the GAN should be able to model. 

The results of the experimental evaluation showed that Sequence GANs are well suited to obtain realistic unobserved event sequences with a relatively small number of unrealistic event sequences. Moreover, the AVATAR generalization scores were compared to existing generalization metrics on the discovered process models. The obtained AVATAR scores on those models were perceived more appropriate than the scores of existing generalization measures based on the ground truth event sequence information.
All experimental results were obtained under ideal working conditions.


\section{Notations}\label{sec:notations}
The notations that are used throughout this paper are based on and consistent with the ones of the original AVATAR publication. Hence, the reader is referred to \cite{theis2020adversarial} for detailed introductions and definitions.

A system is denoted by $S$. 
An event $a \in \mathcal{A}$ describes an instantaneous change of the state of $S$ where $\mathcal{A}$ is the finite set of all possible events. 
The cardinality of a set is denoted by $| \cdot |$. 
An event instance $E$ is a vector and describes the occurrence of a specific $a$ along with its occurrence timestamp and optional additional information.
A trace is a finite and chronologically ordered sequence of event instances. 
A \textit{variant} $v \in \mathcal{V}$ is a sequence of events where $\mathcal{V}$ is the infinite set of all variants. 
A trace maps to exactly one variant.
Whereas an event log is a set of traces, denoted by $\mathcal{L}$, a variant log is a sample of variants denoted by $\mathcal{L}^*$. 
A \textit{unique variant log} is denoted by $\mathcal{L}^+$ and equals to the set of $\mathcal{L}^*$. 
The set of all variants that can be observed during the runtime of $S$ is denoted by $\mathcal{V}_S$.
The functions $\mu(\mathcal{V})$ and $mean(\mathcal{V})$ return the maximum and mean variant lengths of a given set of variants, respectively.

Following the AVATAR methodology, an SGAN architecture is trained on $\mathcal{L}^+$ with a hyperparameter $\beta$. 
The trained GAN can be used to naivey sample variants and is denoted by $GAN_\beta$. 
The number of sampling iterations from $GAN_\beta$ is denoted by $k$. 

When training a GAN, all variants of $\mathcal{L}^+ \subseteq \mathcal{V}_S$ are considered. A subset of variants $\mathcal{V}_u$ might exist such that  $\mathcal{V}_S = (\mathcal{L}^+ \cup \mathcal{V}_u$) and ($\mathcal{L}^+ \cap \mathcal{V}_u) = \emptyset$. 
Ideally, when sampling $k$ times from $GAN_\beta$, it is desired to obtain an estimated set of system variants, i.e. $\hat{\mathcal{V}}_{S}$ that equals to $\mathcal{V}_{S}$. 
How well the GAN performs to reach to this goal is quantified using the true positive ratios described in Equation \ref{eq:tp-S}. 
\begin{equation}\label{eq:tp-S}
\begin{split}
    \displaystyle 
    tp = \frac{|\hat{\mathcal{V}}_S \cap \mathcal{V}_S|}{|\mathcal{V}_S|}, ~
    tp_{u} = \frac{|\hat{\mathcal{V}}_S \cap \mathcal{V}_u|}{|\mathcal{V}_u|}
\end{split}
\end{equation}
$tp$ describes the proportion of realistic variants sampled using $GAN_\beta$ over all possible system variants. 
$tp_{u}$ describes the ratio of sampled variants using $GAN_\beta$ over all unobserved variants. Moreover, the number of unique sampled variants is recorded. 
Ideally, $tp$ and $tp_{u}$ should equal to $1$ while the number of unique sampled variants should equal to $|\mathcal{V}_S|$. 
\begin{equation}\label{eq:score}
    \displaystyle 
    s(tp, tp_u) = \frac{tp + tp_u}{\sqrt{2}}
\end{equation}
The score function described in Equation \ref{eq:score} is used, as proposed in \cite{theis2020adversarial}, to quantify how well the GAN of AVATAR performs. This function addresses the imbalance of $tp$ and $tp_u$. The maximum score value equals $\sqrt{2}$ and is reached at $s(1,1)$.

\section{Problem Statement}\label{sec:problem-statement}
The AVATAR methodology \cite{theis2020adversarial} demonstrated successfully that
Sequence GANs can model $\mathcal{V}_u$ which builds a foundation to measure the generalization of process models.
The evaluation setup of AVATAR consisted of a 70/30 split ratio of $\mathcal{V}_S$ to obtain $\mathcal{L}^+$ and $\mathcal{V}_u$ and a sampling hyperparameter $k$ that was set to $10,000$ for each of the 15 ground truth systems. 
This setup raises multiple research questions, including the following.

\underline{RQ1}: \textit{Is the hyperparameter $k$ with a value of $10,000$ optimally defined and is there a relationship between $k$ and the GAN performance of AVATAR?}
The hyperparameter $k$ describes the number of variants that are drawn naively from the trained Sequence GAN without leveraging the Metropolis-Hastings algorithm. 
Whereas \cite{theis2020adversarial} states that preliminary results showed that setting $k$ to the value of $10,000$ is a good choice, a proven justification for this value is missing. 
Moreover, it remains unclear if there a relationship between $k$ and the performance of the GAN of AVATAR exists. Therefore, this paper experimentally assesses the performance of the GANs with multiple multiple values for $k$ to validate the statement made in the original publication and to investigate the relationship between the system, $k$, and the GAN performance to model $\mathcal{V}_S$.

\underline{RQ2}: \textit{How does the size of $\mathcal{L}^+$ relate to the performance of modeling $\mathcal{V}_S$?} 
The AVATAR methodology has been evaluated using a constant 70/30 split ratio of $\mathcal{V}_S$ to obtain $\mathcal{L}^+$ and $\mathcal{V}_u$ across all used ground truth systems. 
However, it remains unclear how the GAN of AVATAR performs if less information of a system is given. 
In real-world scenarios, an exact 70\% split of all possible variants of a system is usually unrealistic. 
The ratio of variants contained in $\mathcal{L}^+$ to all variants in $\mathcal{V}_S$ can be guessed at its best. 
Hence, this paper experimentally assesses the performance of the GANs of AVATAR at different split ratios to investigate the working conditions of AVATAR when the given event log size is limited.

\underline{RQ3}: \textit{Are the GANs of AVATAR sensitive to biased variant logs?} 
The GANs of AVATAR have been evaluated using a random and unbiased split of $\mathcal{V}_S$. In real-world scenarios though, $\mathcal{L}^+$ might be biased due to a limited observation duration of the system or adverse environmental situations. Whereas research has been conducted on the impact of biased event logs on process discovery algorithms \cite{sani2021impact}, it remains unclear how the GANs of AVATAR perform when being trained on a biased set of variants. Bias can be expressed e.g. in terms of variant lengths. In this paper, preliminary experiments are performed to investigate if the performance of the GANs are affected when being trained on specific biased variant logs.

\begin{figure*}[b!]
  \begin{center}
    \includegraphics[width=\textwidth]{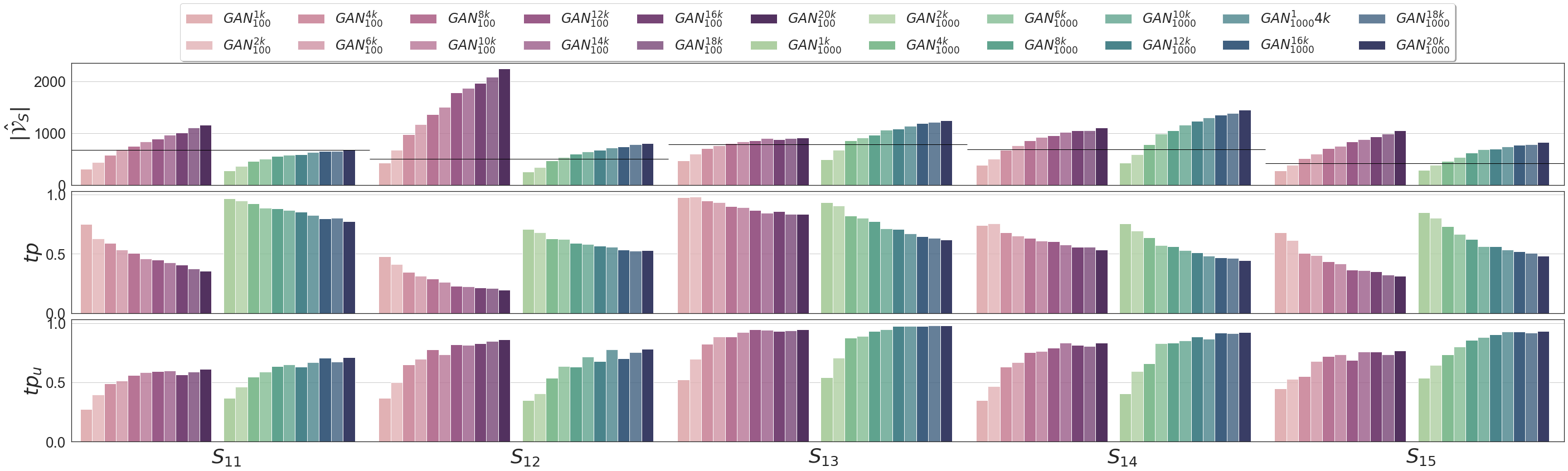}
  \end{center}
  \caption{Obtained results for multiple values of $k$. The gray line in the top graph represents $|V_s|$.}
  \label{fig:exp1_comp}
\end{figure*}

\section{Experimental Setup}\label{sec:exp}
This section describes the experimental setups. The corresponding source code including the obtained results are publicly available on Github\footnote{https://github.com/ProminentLab/AVATAR}.

\subsection{Sampling Hyperparameter}\label{sec:rq1}
To investigate the relationship between $k$ and the performance of the Sequence GANs of AVATAR (\textit{RQ1}), multiple values for $k$ are investigated. 
Specifically, $k$ is set to $1,000$, $2,000$, $4,000$, $6,000$, up to $20,000$, with an increment of $2,000$ each. 
This includes the originally proposed $k=10,000$ value. 
These specific values are chosen such that performance changes can be observed when increasing and decreasing the proposed value of $k$. 
It is expected that the performance of the GANs decreases with a very small value, such as $k=1,000$, but it remains unclear if the performance increases with an increased value of $k$.
It is not expected that a granularity finer than $2,000$ will unveil significant differences.

Training and sampling of the Sequence GANs is performed on the five highly complex Petri net ground truth system that were also used to evaluate the AVATAR methodology according to the original publication. 
These systems are denoted as $S_{11-15}$ and correspond to Systems 11-15 in \cite{theis2020adversarial}.
For each of the five systems, two Sequence GANs are trained with $\beta = 100$ and $\beta = 1000$, respectively. 
These GANs are trained using a random 70\% split of $\mathcal{V}_S$ which corresponds to $\mathcal{L}^+$. 
The remaining 30\% results in $\mathcal{V}_u$ and are used to evaluate the performance of the Sequence GAN to approximate the unobserved system variants, as in the original publication. This is called a \textit{70/30 split ratio}.
The setup results in ten different Sequence GAN models and, due to to $11$ different values for $k$, in a total of 110 observation values for evaluation.

\subsection{Variant Log Size}\label{sec:rq2}
To investigate the performance of the GANs of AVATAR when limited variant log sizes are given (\textit{RQ2}), two Sequence GANs per system are trained with different split ratios compared to the 70/30 ratio of the original evaluation. 
In this setup, the 70/30 split ratio is used as a baseline for comparison.
Moreover, experiments are performed using 10/90, 20/80, 30/70, 40/60, 50/50, and 60/40 split ratios. 
It is expected that the performance of the GANs in modeling $\mathbf{V}_u$ decreases with smaller $|\mathcal{L}^+|$ values.
As before, the systems $S_{11-15}$ are used for experimental evaluation due to their realistic complexity.
The Sequence GANs are trained with $\beta = 100$ and $\beta = 1000$ to be consistent with the original AVATAR work.
This results in 140 Sequence GANs for evaluation. 
Variants are generated from the Sequence GANs using the originally proposed $k=10,000$ value.
Statistics on the variant sets that are used for this experiment are shown in Table \ref{tab:exp2_variantstats}.

\begin{table}[h]
\centering
\resizebox{250pt}{!}{%
\begin{tabular}{|c|c|c|c|c|c|c|c|c|}
\hline
\multicolumn{1}{|l|}{}         & \multicolumn{1}{l|}{} & \multicolumn{3}{c|}{\textbf{$S$}}                                           & \multicolumn{2}{c|}{\textbf{$\mathcal{L}^+$}} & \multicolumn{2}{c|}{\textbf{$\mathcal{V}_u$}} \\ \hline
\textbf{Identifier}                   & \textbf{Setup}           & \textbf{$|\mathcal{A}|$} & \textbf{$|\mathcal{V}_S|$} & \textbf{$\mu(\mathcal{V}_S)$} & \textbf{$|\mathcal{L}^+|$}   & \textbf{$mean(\mathcal{L}^+)$}  & \textbf{$|\mathcal{V}_u|$}   & \textbf{$mean(\mathcal{V}_u)$}   \\ \hline
\multirow{7}{*}{pb\_system\_1\_5} & baseline 70/30           & \multirow{7}{*}{14}       & \multirow{7}{*}{680} & \multirow{7}{*}{18}         & 476                    & 13.51                 & 204                    & 13.67                  \\ \cline{2-2} \cline{6-9} 
                                  & 10/90                    &                           &                      &                             & 68                     & 13.26                 & 612                    & 13.59                  \\ \cline{2-2} \cline{6-9} 
                                  & 20/80                    &                           &                      &                             & 136                    & 13.51                 & 544                    & 13.57                  \\ \cline{2-2} \cline{6-9} 
                                  & 30/70                    &                           &                      &                             & 204                    & 13.94                 & 476                    & 13.40                  \\ \cline{2-2} \cline{6-9} 
                                  & 40/60                    &                           &                      &                             & 272                    & 13.46                 & 408                    & 13.63                  \\ \cline{2-2} \cline{6-9} 
                                  & 50/50                    &                           &                      &                             & 340                    & 13.65                 & 340                    & 13.47                  \\ \cline{2-2} \cline{6-9} 
                                  & 40/60                    &                           &                      &                             & 408                    & 13.56                 & 272                    & 13.56                  \\ \hline
\multirow{7}{*}{pb\_system\_2\_4} & baseline 70/30           & \multirow{7}{*}{15}       & \multirow{7}{*}{507} & \multirow{7}{*}{43}         & 355                    & 30.98                 & 152                    & 31.30                  \\ \cline{2-2} \cline{6-9} 
                                  & 10/90                    &                           &                      &                             & 51                     & 31.14                 & 456                    & 31.07                  \\ \cline{2-2} \cline{6-9} 
                                  & 20/80                    &                           &                      &                             & 102                    & 31.04                 & 405                    & 31.08                  \\ \cline{2-2} \cline{6-9} 
                                  & 30/70                    &                           &                      &                             & 152                    & 31.43                 & 355                    & 30.92                  \\ \cline{2-2} \cline{6-9} 
                                  & 40/60                    &                           &                      &                             & 203                    & 30.80                 & 304                    & 31.25                  \\ \cline{2-2} \cline{6-9} 
                                  & 50/50                    &                           &                      &                             & 254                    & 31.48                 & 253                    & 30.67                  \\ \cline{2-2} \cline{6-9} 
                                  & 40/60                    &                           &                      &                             & 304                    & 30.71                 & 203                    & 31.62                  \\ \hline
\multirow{7}{*}{pb\_system\_3\_6} & baseline 70/30           & \multirow{7}{*}{10}       & \multirow{7}{*}{780} & \multirow{7}{*}{16}         & 546                    & 10.26                 & 234                    & 10.28                  \\ \cline{2-2} \cline{6-9} 
                                  & 10/90                    &                           &                      &                             & 78                     & 10.51                 & 702                    & 10.24                  \\ \cline{2-2} \cline{6-9} 
                                  & 20/80                    &                           &                      &                             & 156                    & 10.26                 & 624                    & 10.27                  \\ \cline{2-2} \cline{6-9} 
                                  & 30/70                    &                           &                      &                             & 234                    & 10.27                 & 546                    & 10.26                  \\ \cline{2-2} \cline{6-9} 
                                  & 40/60                    &                           &                      &                             & 312                    & 10.28                 & 468                    & 10.26                  \\ \cline{2-2} \cline{6-9} 
                                  & 50/50                    &                           &                      &                             & 390                    & 10.17                 & 390                    & 10.36                  \\ \cline{2-2} \cline{6-9} 
                                  & 40/60                    &                           &                      &                             & 468                    & 10.28                 & 312                    & 10.25                  \\ \hline
\multirow{7}{*}{pb\_system\_4\_1} & baseline 70/30           & \multirow{7}{*}{15}       & \multirow{7}{*}{688} & \multirow{7}{*}{28}         & 481                    & 21.64                 & 207                    & 21.98                  \\ \cline{2-2} \cline{6-9} 
                                  & 10/90                    &                           &                      &                             & 69                     & 21.58                 & 619                    & 21.76                  \\ \cline{2-2} \cline{6-9} 
                                  & 20/80                    &                           &                      &                             & 138                    & 22.02                 & 550                    & 21.67                  \\ \cline{2-2} \cline{6-9} 
                                  & 30/70                    &                           &                      &                             & 207                    & 22.05                 & 481                    & 21.60                  \\ \cline{2-2} \cline{6-9} 
                                  & 40/60                    &                           &                      &                             & 275                    & 21.69                 & 413                    & 21.77                  \\ \cline{2-2} \cline{6-9} 
                                  & 50/50                    &                           &                      &                             & 344                    & 21.68                 & 344                    & 21.80                  \\ \cline{2-2} \cline{6-9} 
                                  & 40/60                    &                           &                      &                             & 413                    & 21.80                 & 275                    & 21.65                  \\ \hline
\multirow{7}{*}{pb\_system\_5\_3} & baseline 70/30           & \multirow{7}{*}{14}       & \multirow{7}{*}{415} & \multirow{7}{*}{21}         & 290                    & 14.65                 & 125                    & 14.88                  \\ \cline{2-2} \cline{6-9} 
                                  & 10/90                    &                           &                      &                             & 42                     & 14.40                 & 373                    & 14.75                  \\ \cline{2-2} \cline{6-9} 
                                  & 20/80                    &                           &                      &                             & 83                     & 15.42                 & 332                    & 14.54                  \\ \cline{2-2} \cline{6-9} 
                                  & 30/70                    &                           &                      &                             & 125                    & 14.49                 & 290                    & 14.82                  \\ \cline{2-2} \cline{6-9} 
                                  & 40/60                    &                           &                      &                             & 166                    & 14.37                 & 249                    & 14.95                  \\ \cline{2-2} \cline{6-9} 
                                  & 50/50                    &                           &                      &                             & 208                    & 14.43                 & 207                    & 15.01                  \\ \cline{2-2} \cline{6-9} 
                                  & 40/60                    &                           &                      &                             & 249                    & 14.66                 & 166                    & 14.80                  \\ \hline
\end{tabular}%
}
\caption{Variant set statistics for limited variant log size experiment}
\label{tab:exp2_variantstats}
\end{table}

\subsection{Biased Variant Logs}\label{sec:rq3}
This experiment investigates the performance of the GANs of AVATAR in detecting $\mathcal{V}_u$ when being trained on a biased $\mathcal{L}^+$ to provide an answer to \textit{RQ3}. 
Bias is expressed using the length of variants. 
The baseline is obtained using a random and unbiased 70/30 split ratio on $\mathcal{V}_S$ such that $mean(\mathcal{L}^+)$ and $mean(\mathcal{V}_u)$ are almost equal. 
Four bias setups are defined and denoted by $b1$ to $b4$.

The first bias setup $b1$ is defined such that $\mathcal{L}^+$ contains the shortest 70\% of $\mathcal{V}_s$ and $\mathcal{V}_u$ contains the remaining 30\%. This means that a Sequence GAN is trained on short variants only and is supposed to generalize long variants, too.

Similarly, the second bias setup $b2$ is defined such that $\mathcal{L}^+$ contains the longest 70\% of $\mathcal{V}_s$ and $\mathcal{V}_u$ contains the remaining variants. In this case, a Sequence GAN is trained on long variants and is supposed to generalize short variants.

The third and fourth setups $b3$ and $b4$ are leaky variations of $b1$ and $b2$, respectively. For both setups, 20\% of the variants in $\mathcal{V}_u$ are randomly exchanged with a randomly chosen variant from $\mathcal{L}^+$. This means that the corresponding Sequence GAN is not trained on strictly short or strictly long variants. However, bias in terms of the lengths of variants contained in $\mathcal{L}^+$ and $\mathcal{V}_u$ persists. 

For all setups $b1$ to $b4$, the longest possible variant of a corresponding system is contained in $\mathcal{L}^+$ rather than $\mathcal{V}_u$. 
This is required to satisfy the assumption that the maximum possible system variant length is known to train a GAN of AVATAR. Therefore, at least one variant with a length equal to $\mu(\mathcal{V}_S)$ must be known.

Like before, two Sequence GANs are trained with $\beta=100$ and $\beta=1000$, respectively, for each of the systems $S_{11-15}$ and each setup. Consequently, the total number of Sequence GAN models under investigation equals 50. 
Table \ref{tab:exp3_variantstats} shows the variant set statistics for each system and setup under investigation.

\begin{table}[]
\centering
\resizebox{250pt}{!}{%
\begin{tabular}{|c|c|c|c|c|c|c|c|c|}
\hline
\multicolumn{1}{|l|}{}         & \multicolumn{1}{l|}{} & \multicolumn{3}{c|}{\textbf{$S$}}                                           & \multicolumn{2}{c|}{\textbf{$\mathcal{L}^+$}} & \multicolumn{2}{c|}{\textbf{$\mathcal{V}_u$}} \\ \hline
\textbf{Identifier}                   & \textbf{Setup}           & \textbf{$|\mathcal{A}|$} & \textbf{$|\mathcal{V}_S|$} & \textbf{$\mu(\mathcal{V}_S)$} & \textbf{$|\mathcal{L}^+|$}   & \textbf{$mean(\mathcal{L}^+)$}  & \textbf{$|\mathcal{V}_u|$}   & \textbf{$mean(\mathcal{V}_u)$}   \\ \hline
\multirow{5}{*}{$S_{11}$} & baseline                 & 14                        & 680                  & 18                          & 476                    & 13.51                 & 204                    & 13.67                  \\ \cline{2-9} 
                                  & $b1$                       & 14                        & 680                  & 18                          & 477                    & 12.72                 & 203                    & 15.54                  \\ \cline{2-9} 
                                  & $b2$                       & 14                        & 680                  & 18                          & 476                    & 14.62                 & 204                    & 11.08                  \\ \cline{2-9} 
                                  & $b3$                       & 14                        & 680                  & 18                          & 477                    & 13.03                 & 203                    & 14.81                  \\ \cline{2-9} 
                                  & $b4$                       & 14                        & 680                  & 18                          & 476                    & 14.27                 & 204                    & 11.90                  \\ \hline
\multirow{5}{*}{$S_{12}$} & baseline                 & 15                        & 507                  & 43                          & 355                    & 30.98                 & 152                    & 31.30                  \\ \cline{2-9} 
                                  & $b1$                       & 15                        & 507                  & 43                          & 356                    & 27.96                 & 151                    & 38.41                  \\ \cline{2-9} 
                                  & $b2$                       & 15                        & 507                  & 43                          & 355                    & 34.28                 & 152                    & 23.59                  \\ \cline{2-9} 
                                  & $b3$                       & 15                        & 507                  & 43                          & 356                    & 29.01                 & 151                    & 35.93                  \\ \cline{2-9} 
                                  & $b4$                       & 15                        & 507                  & 43                          & 355                    & 33.32                 & 152                    & 25.82                  \\ \hline
\multirow{5}{*}{$S_{13}$} & baseline                 & 10                        & 780                  & 16                          & 546                    & 10.26                 & 234                    & 10.28                  \\ \cline{2-9} 
                                  & $b1$                       & 10                        & 780                  & 16                          & 547                    & 8.92                  & 233                    & 13.43                  \\ \cline{2-9} 
                                  & $b2$                       & 10                        & 780                  & 16                          & 546                    & 11.61                 & 234                    & 7.13                   \\ \cline{2-9} 
                                  & $b3$                       & 10                        & 780                  & 16                          & 547                    & 9.44                  & 233                    & 12.20                  \\ \cline{2-9} 
                                  & $b4$                       & 10                        & 780                  & 16                          & 546                    & 11.15                 & 234                    & 8.21                   \\ \hline
\multirow{5}{*}{$S_{14}$} & baseline                 & 15                        & 688                  & 28                          & 481                    & 21.64                 & 207                    & 21.98                  \\ \cline{2-9} 
                                  & $b1$                       & 15                        & 688                  & 28                          & 482                    & 19.91                 & 206                    & 26.01                  \\ \cline{2-9} 
                                  & $b2$                       & 15                        & 688                  & 28                          & 481                    & 23.74                 & 207                    & 17.05                  \\ \cline{2-9} 
                                  & $b3$                       & 15                        & 688                  & 28                          & 482                    & 20.64                 & 206                    & 24.31                  \\ \cline{2-9} 
                                  & $b4$                       & 15                        & 688                  & 28                          & 481                    & 23.17                 & 207                    & 18.40                  \\ \hline
\multirow{5}{*}{$S_{15}$} & baseline                 & 14                        & 415                  & 21                          & 290                    & 14.65                 & 125                    & 14.88                  \\ \cline{2-9} 
                                  & $b1$                       & 14                        & 415                  & 21                          & 291                    & 12.94                 & 124                    & 18.90                  \\ \cline{2-9} 
                                  & $b2$                       & 14                        & 415                  & 21                          & 290                    & 16.80                 & 125                    & 9.84                   \\ \cline{2-9} 
                                  & $b3$                       & 14                        & 415                  & 21                          & 291                    & 13.81                 & 124                    & 16.86                  \\ \cline{2-9} 
                                  & $b4$                       & 14                        & 415                  & 21                          & 290                    & 16.05                 & 125                    & 11.59                  \\ \hline
\end{tabular}%
}
\caption{Variant set statistics for the biased variant log experiment}
\label{tab:exp3_variantstats}
\end{table}

\section{Results}\label{sec:results}	
This section describes and discusses the obtained results of the experiments that are defined above.

\begin{figure*}[ht!]
  \begin{center}
    \includegraphics[width=\textwidth]{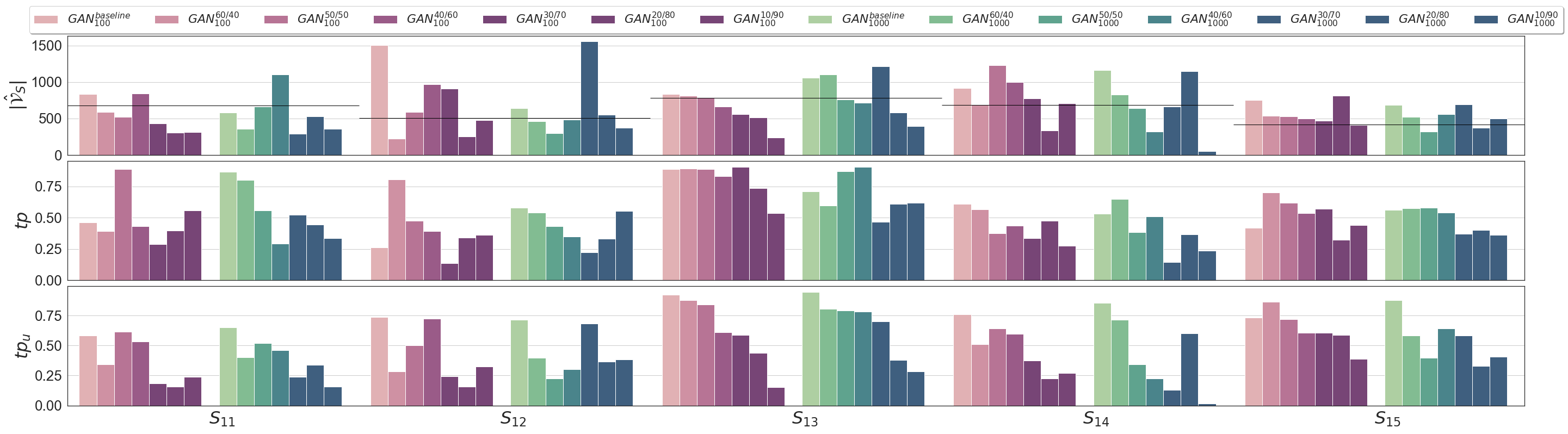}
  \end{center}
  \caption{Obtained results for different split setups compared to the baseline. The gray line in the top graph represents $|V_s|$.}
  \label{fig:exp2_comp}
\end{figure*}

\subsection{Sampling Hyperparameter Results}\label{sec:results-rq1}
The obtained results are visualized in Figure \ref{fig:exp1_comp}. For $S_{11}$ and $GAN_{100}$, it can be observed that the number of approximated system variants increases with the value of $k$. This GAN setup is closest to the desired $|\mathcal{V}_S|$ value when using $k=8,000$. In the meantime, the $tp$ ratio decreases with an increasing value of $k$. With an increasing value of $k$, the $tp_u$ ratio converges to $0.6$. 
Similar behavior can be observed for the Sequence GANs for $S_{12}$. However, with $k=2,000$, $\hat{\mathcal{V}}_S$ already exceeds the desired value of $\mathcal{V}_S$. 
Accordingly, $tp$ decreases and $tp_u$ converges with increasing $k$ to about $0.8$. The overestimation of variants can be explained by the complexity of the underlying system. 
$S_{12}$ is the most complex system with $15$ unique events and a maximum variant length of $43$. The second most complex system is $S_{14}$ with a much smaller maximum variant length. 
Accordingly, the Sequence GANs of $S_{14}$ are better in approximating $\mathcal{V}_S$ compared to the ones of $S_{12}$. 
Systems $S_{13-15}$ perform similarly to $S_{11}$ with an optimal variant number approximation around $k=10000$. 
The $tp_u$ ratios usually seem to converge around $0.7$ and $0.9$.

The results look similar for GANs with $\beta=1,000$. 
In general, $\hat{\mathcal{V}}_S$ is overestimated with an increasing value of $k$ and when using a value that is larger than $10,000$.
Only for $S_{11}$, the corresponding Sequence GAN underestimates the number of samples when using any of the considered values for $k$. 
However, for $k=20,000$, $GAN_{1000}$ almost perfectly estimates $|\mathcal{V}_S|$ with a decently high $tp$ and $tp_u$ ratio.
In general, the $tp$ ratio reduces with a more gentle slope compared to $GAN_{100}$ while $tp_u$ converges to a fixed value similar to $GAN_{100}$. The $tp_u$ convergence value lies between $0.75$ and $1.0$

\begin{figure}[h]
  \begin{center}
    \includegraphics[width=240pt]{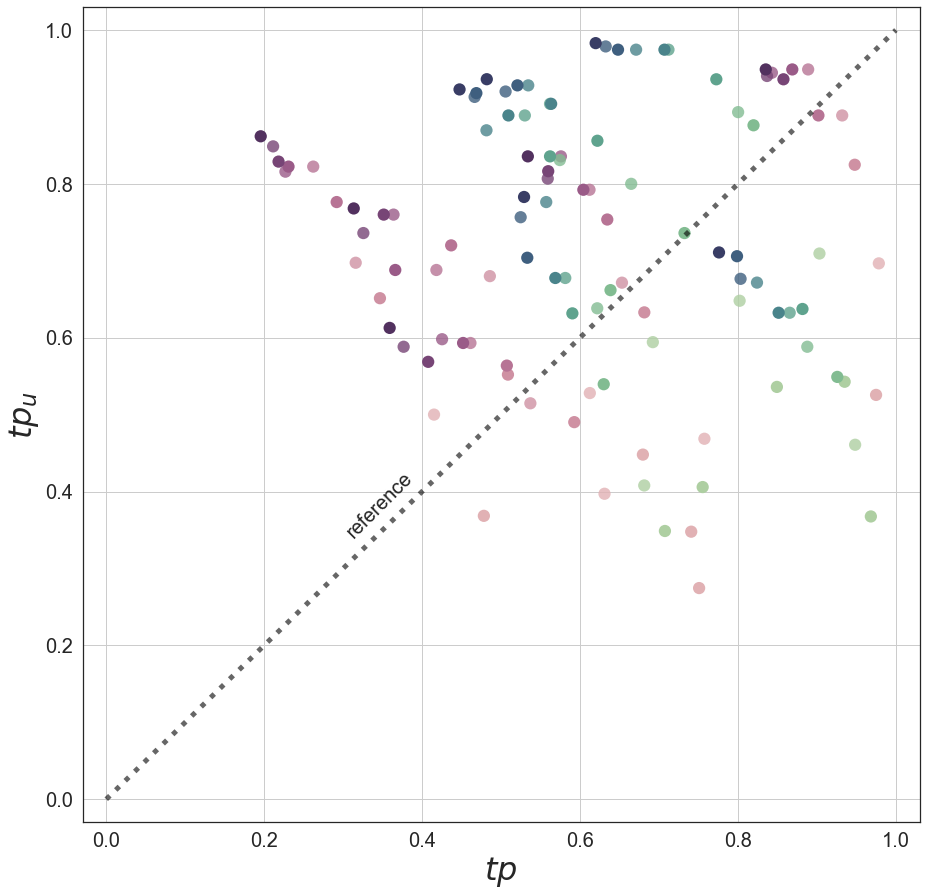}
  \end{center}
  \caption{2D visualization of the scores for the obtained results at multiple values of $k$.}
  \label{fig:exp1_2d}
\end{figure}

Figure \ref{fig:exp1_2d} shows the corresponding scores $s$ to address the tradeoff between $tp$ and $tp_u$. It can be observed that for smaller values of $k$, the resulting score lies mostly in the upper right area, but below the reference line. In comparison, for larger $k$ values, the resulting score lies above the reference line. 
Generally, the best scores are obtained for intermediate values of $k$. This indicates a quadratic relationship between $k$ and $s$. 

Since it can be observed that the performance of the GANs on more complex systems, such as $S_{12}$, can be weaker, a linear regression model is fit using the features $k$, $\mu(\mathcal{V}_S)$, and $|\mathcal{V}_S|$ to model the resulting scoring value for $s$. With linear features, this leads to a and $R^2$ value of $1.4\%$ indicating a bad fit. With the corresponding quadratic features, the $R^2$ score improves to $40\%$. This underscores the quadratic relationship that could be an initial step to develop a rule-of-thumb to select an individual and optimized value for $k$. The required values for $|\mathcal{V}_S|$, $\mu(\mathcal{V}_S)$ and a desired minimum score value $s$ can be guessed by using expert knowledge. 
    
The median value of the best obtained scores for the Sequence GAN models under consideration equals $10,000$. This validates the general suitability of $k=10,000$ as proposed in \cite{theis2020adversarial} and answers \textit{RQ1}.

\subsection{Variant Log Size Results}\label{sec:results-rq2}

\begin{figure*}[ht!]
  \begin{center}
    \includegraphics[width=\textwidth]{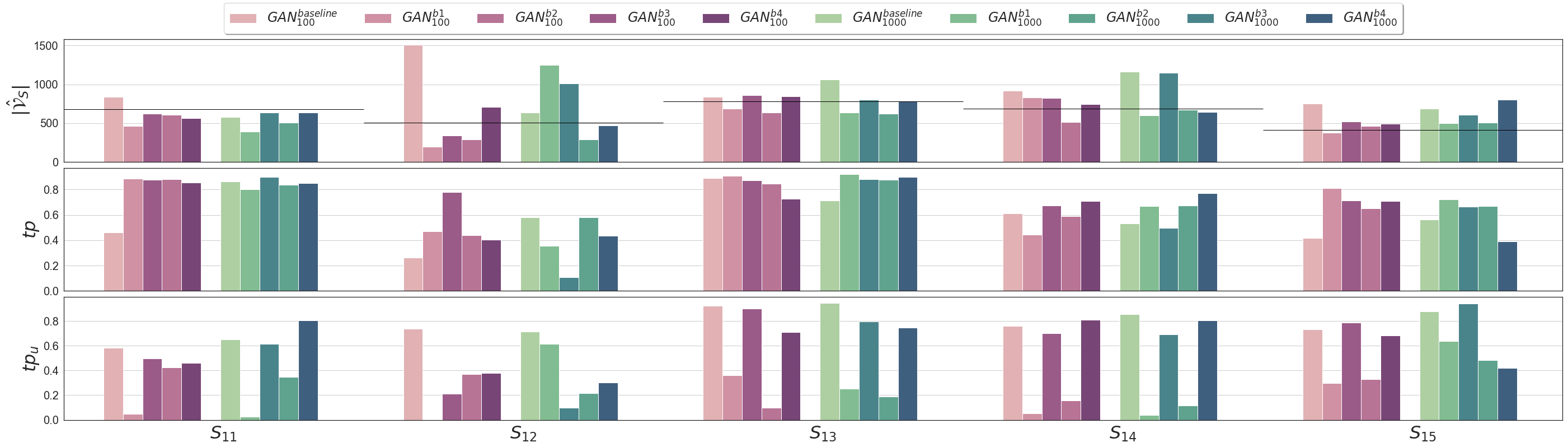}
  \end{center}
  \caption{Obtained results for biased setups compared to the baseline. The gray line in the top graph corresponds to $|V_s|$.}
  \label{fig:exp3_comp}
\end{figure*}

The obtained results are visualized in Figure \ref{fig:exp2_comp}. For the GANs that were trained using $\beta=100$, it can be generally noted that fewer unique variants are sampled with decreasing sizes of $\mathcal{L}^+$. 
At the same time, it can also be observed that $tp$ and $tp_u$ generally tend to decrease. 
A similar, but less significant behavior can be observed for the Sequence GANs that are trained using $\beta=1000$. This confirms the expectations. 

\begin{figure}[hb]
  \begin{center}
    \includegraphics[width=240pt]{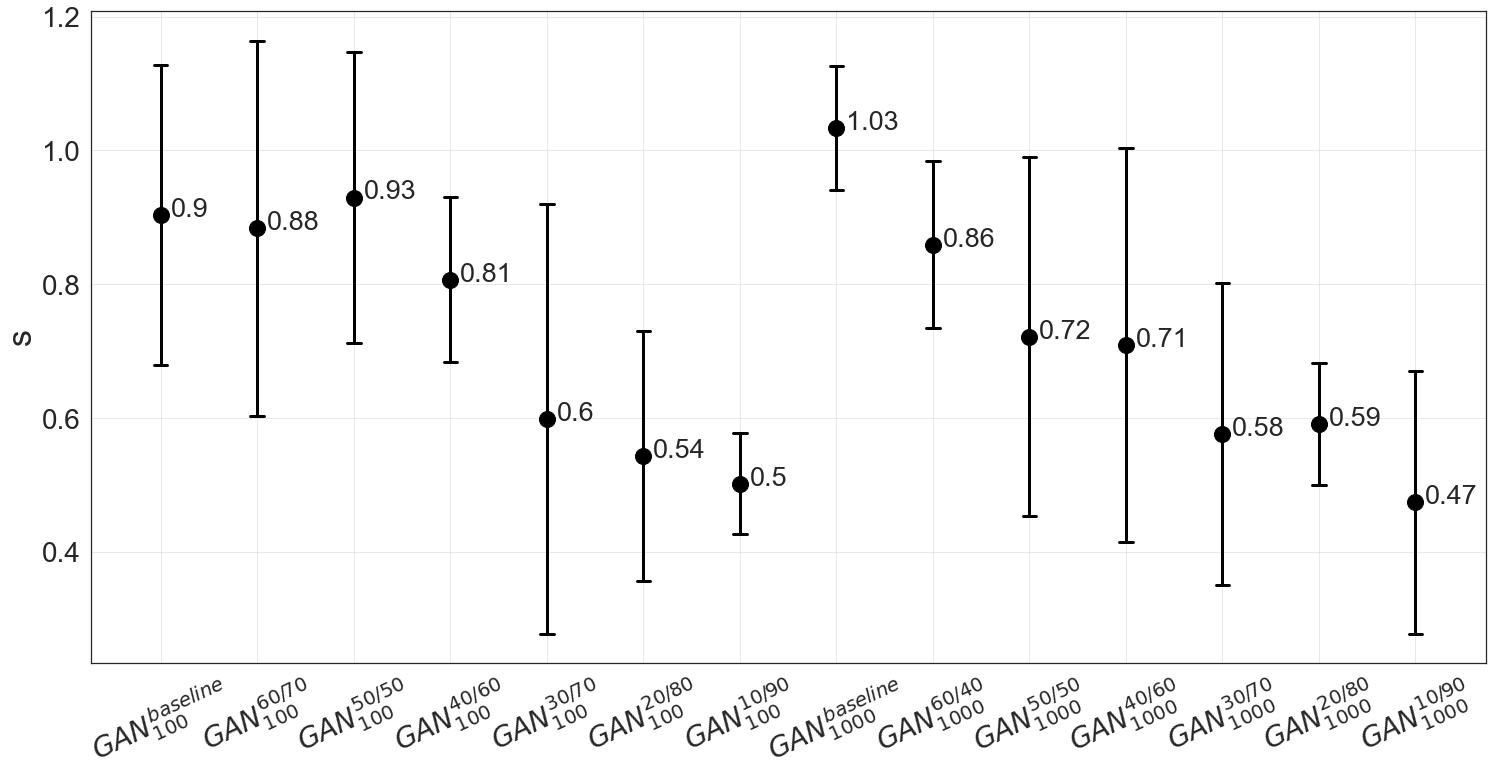}
  \end{center}
  \caption{90\% CIs of the mean scores $s$ for each Sequence GAN setup of different $\mathcal{L}^+$ sizes over all systems $S_{11-15}$}
  \label{fig:exp2_ci}
\end{figure}

The same trend can be observed when visualizing the 90\% confidence intervals (CI) of the obtained scores $s$ for each Sequence GAN and variant log size setup over all systems, as visualized in Figure \ref{fig:exp2_ci}. Whereas this visualization cannot provide statistical proof due to the small sample size, it nonetheless shows the decreasing trend satisfyingly. 
Moreover, since the CIs for a 10/90 split ratio and the baseline 70/30 split ratio for both Sequence GAN setups are non-overlapping, it can be concluded that a 10/90 split ratio performs statistically poorer than a 70/30 split ratio with $90\%$ confidence.

To provide an answer to \textit{RQ2}, the GAN performance decreases with a decreasing number of observed variants contained in $\mathcal{L}^+$ relative to $|\mathcal{V}_S|$. 
These experiments prove that the Sequence GANs of AVATAR trained with a 70/30 split ratio perform statistically significantly better compared to a 10/90 split ratio.
For $GAN_{1000}$, the experiments show that a 70/30 split ratio leads to statistically significantly better performance compared to 30/70, 20/80, and 10/90 split ratios.
However, it needs to be highlighted that further experiments with a larger sample size are required to provide a stronger validation of the statements. 

\subsection{Biased Variant Log Results}\label{sec:results-rq3}
The results of the biased variant log experiments are visualized in Figure \ref{fig:exp3_comp}. It can be observed that for all systems, the Sequence GAN using $\beta=100$ on the biased setup $b1$ performs poorly. However, when training using $\beta=1000$, the performance seems to be increasing. It can be noticed that the $\beta$ parameter seems to have an impact on the performance when $\mathcal{L}^+$ is biased. However, the details of the impact remain unclear. 
Overall, the Sequence GANs trained with $\beta=1000$ seem to perform better in general based on the result visualization in Figure \ref{fig:exp3_comp}. 

\begin{figure}[h]
  \begin{center}
    \includegraphics[width=240pt]{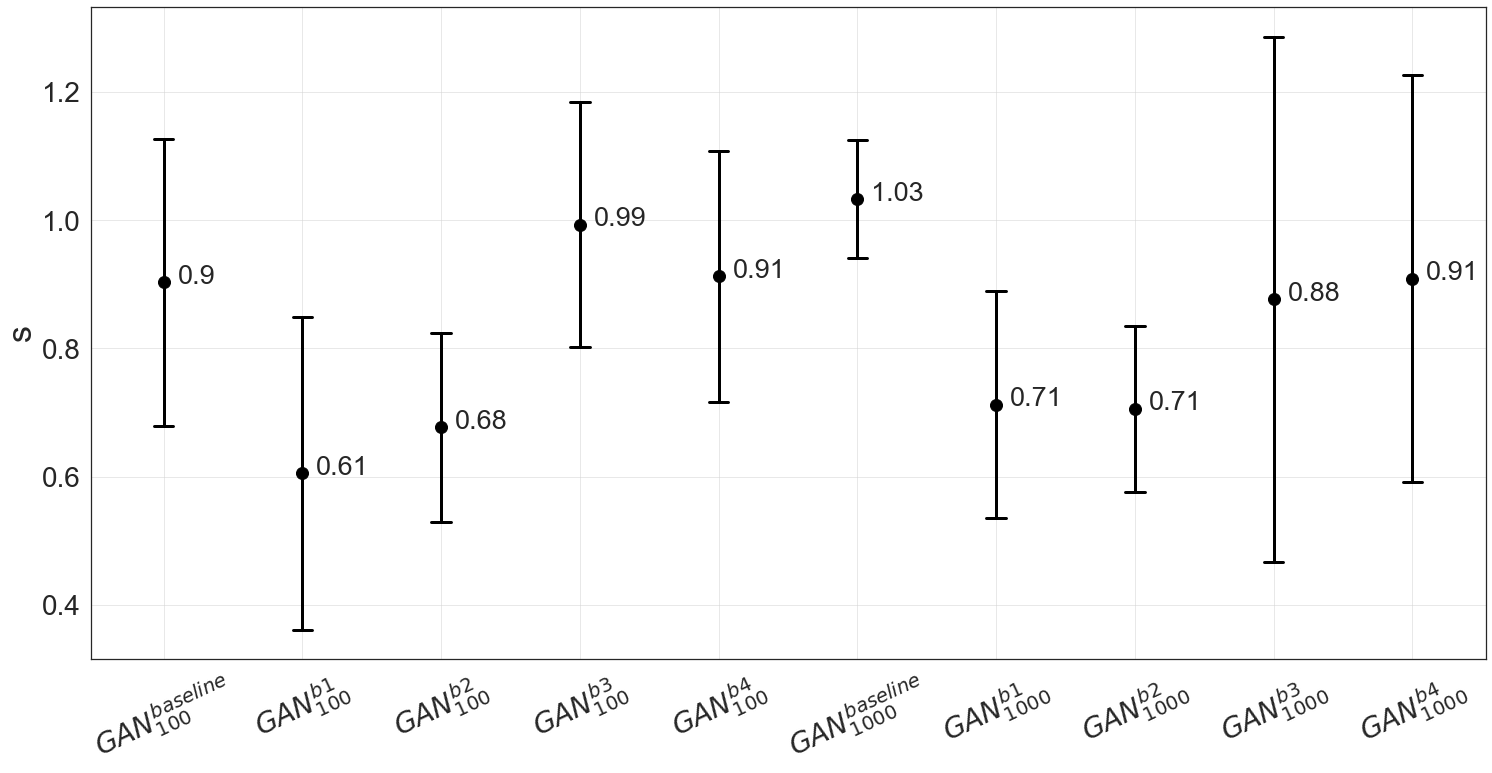}
  \end{center}
  \caption{90\% CIs of the mean scores $s$ for each biased $\mathcal{L}^+$ and baseline Sequence GAN setup over all systems $S_{11-15}$}
  \label{fig:exp3_ci}
\end{figure}

Furthermore, the GAN performance of AVATAR seems to increase when $\mathcal{L}^+$ is less restrictively biased, i.e. with the setups $b3$ and $b4$ compared to $b1$ and $b2$, respectively. 
Additionally, $b2$ seems to perform better than $b1$, and $b4$ performs better than $b3$. This indicates that less bias leads to better performance.
The same can be observed when visualizing the $90\%$ CIs of the scores $s$ per Sequence GAN setup over all systems, as visualized in Figure \ref{fig:exp3_ci}. 
Comments on the statistical significance of each CI cannot be made due to the small sample size. However, the CI mean values indicate the trend that is described above. 
The baseline Sequence GANs are overall the best-performing models. When introducing leaky bias with the setups $b3$ and $b4$, the performance reduces on average. Strict bias, such as with setups $b1$ and $b2$, leads to a further decrease of performance in unveiling $\mathcal{V}_S$. 
The large confidence intervals for the Sequence GANs trained using $\beta=1000$ and using the setups $b3$ and $b4$ can be either a randomness artifact or a sign that the $\beta$ hyperparameter can accommodate for non-strict bias in specific situations. 

To answer to \textit{RQ3}, the GANs of AVATAR are sensitive to bias and perform with a score value $s$ which decreases proportionally to the significance of present variant length bias in  $\mathcal{L}^+$. Further experiments with a larger sample size of ground truth systems are anticipated to provide statistical evidence and insights on the potential impact of the $\beta$ hyperparameter to accommodate for bias. 

\section{Conclusion}\label{sec:conclusion}
This paper investigated the performance of the GANs of the AVATAR method under non-ideal conditions to raise awareness about the working conditions of the methodology to quantify process model generalization. 
Specifically, the sampling hyperparameter $k$ and the sensitivity of the Sequence GANs of AVATAR to limited and biased event logs have been analyzed. Answers to three RQs are provided based on the obtained results.

Regarding \textit{RQ1}, the experiments have shown that $k=10,000$ is generally a good choice. However, an individual value for $k$ is required depending on the underlying system complexity to fine-tune the GAN performance. 
Linear regression with quadratic features indicated a good fit to estimate an optimized value for $k$ given the desired performance score $s$, the total number of system variants, and the maximum variant length of the underlying system. 
For \textit{RQ2}, the GAN performance in modeling $\mathcal{V}_S$ generally tends to decrease when fewer variants of the system are contained in $\mathcal{L}^+$. 
Finally, the GANs of AVATAR seem to be sensitive towards biased variant logs, as an answer to \textit{RQ3}. The performance of the underlying Sequence GANs decreases the more significant the bias in $\mathcal{L}^+$ is. Moreover, the experimental results show the potential that the Sequence GAN hyperparameter $\beta$ might be able to accommodate for bias in specific situations. 

While the experimental results unequivocally highlight certain conditions of the GANs that need to be considered when applying AVATAR, detailed statistical evidence remains mostly missing due to limited sample sizes. Hence, the results of this paper should raise awareness to the research community and provide the following three research directions.
First, the results of the hyperparameter $k$ investigations motivate future experimental evaluations to derive a rule-of-thumb to select an optimal value $k$. This requires an experimental evaluation using a large set of different ground truth systems to derive a robust rule-of-thumb.
Second, a larger set of experiments need to be conducted to investigate the required variant log size to train a converging GAN such that AVATAR can be applied confidently.
Third, the bias sensitivity of the GANs of AVATAR needs to be investigated with a larger set of ground truth systems and with different $\beta$ hyperparameter values to unveil a potential relationship between $\beta$ and the GAN sensitivity towards bias.

\bibliographystyle{IEEEtran}
\bibliography{IEEEabrv,references.bib}
\end{document}